\documentstyle[12pt]{article}
\title{Beliefs and Probability in Bacchus' l.p. Logic: A~3-Valued    
          Logic Solution to Apparent Counter-intuition }

\author{Mieczyslaw A. Klopotek\\
        \em Institute of Computer Science  Polish Academy of Sciences \\
        \em  Warsaw,  Poland}

\date{}

\begin{document}
\setlength{\unitlength}{1cm}

\maketitle
\thispagestyle{empty}

\begin {abstract}
Fundamental discrepancy between first order logic and statistical inference 
(global versus local
properties of universe) is shown to be the obstacle for integration of logic and
probability 
in L.p. logic of Bacchus. To overcome the counterintuitiveness of L.p.
behaviour, a 3-valued logic is proposed.
\end {abstract}

\section{Introduction}

The paper of Bacchus$^{1}$   aims  at  painless  integration  of  two 
paradigms of human reasoning,  that is 1) first order  logics  and 
2) statistical inference. (see also$^{8}$ )  in such a way  as  to 
avoid all the contradictions emerging in previous approaches.

     Nonetheless the claim of the current paper is that  also  the 
L.p. logic of  Bacchus$^{1}$   fails to achieve  its  primary  goal  of 
becoming the tool for describing knowledge \& reasoning  in  expert 
systems and other knowledge-based systems. In Section 2 we present 
several simple examples of basic flaws of  this  logic  exploiting 
the  counter  intuitiveness  of  the   L.p.   logic.   Section   3 
demonstrates a more elaborated example pointing at  weaknesses  of 
the L.p. logic.

     As a remedy  we  propose  (in  a  sketchy  way)  a  different 
deduction theory taking into account the gap  between  first-order 
way  of  thinking  (global  treatment  of  domains)  and  that  of 
statistical (experimental) sciences (local treatment of domains). 

\section{Flaws of Theories Criticized by Bacchus and His Solution}
 
     A number of works were concerned  with  representational  and 
inferential issues when probabilities of events were identified as 
degrees  of  belief.  Bacchus  criticized   i.e.   the   following 
approaches:

\noindent
{\bf Approach 1:} (propositional logic) : (see.$^{6,7}$ ) 
Probability of  a 
sentence is the probability of selection of one of those  possible 
worlds wherein this sentence holds. E.g. the 90 \% belief that  the 
famous Tweety flies is  stated  as  $Prob(Flies(Tweety))=x$  with  $x$ 
being greater than 0.9. However, such an approach does not make it 
easy to state that ``Most birds fly''.

\noindent
{\bf Approach 2:}  (first  order  logic)   The   probability   of   the 
expression: $ \forall x.Bird(x) \rightarrow Flies(x)$  be  expressed      as 
$Prob(\forall x.Bird(x) \rightarrow Flies(x))$.
   Following    the    principles    of 
probability  calculus   we   obtain:   
$Prob(\exists \  x.Bird(x) \wedge \neg Flies(x))= 
Prob( \neg (\forall x. Bird(x) \rightarrow Flies(x))) 
=1-Prob(\forall x.Bird(x) \rightarrow Flies(x)))$. 
Hence if $Prob(\forall x.Bird(x) \rightarrow Flies(x))>0.9$, 
 then it should hold that 
$Prob(\exists \  x.Bird(x) \wedge \neg Flies(x)) < 0.1$. 
 However one can imagine  such  a 
set of possible worlds that in most of those worlds most of  birds 
fly and at the same time in most of the  worlds   non-flying 
birds, exist,   that   is   both    
$Prob(\forall x.Bird(x) \rightarrow Flies(x))>0.9$    and 
$Prob(\exists \  x.Bird(x) \wedge \neg Flies(x)) >  0.9  >  0.1$  
 hold  which  means   a 
contradiction. 

\noindent
{\bf Approach 3:}  Cheeseman$^{5}$   proposed that 
the above  statements  be 
meta-expression with conditional probability of the type:
$$      \forall x.Prob[Flies(x) \mid Bird(x)] > 0.9$$
However, this representation cannot be  treated  as  a  method  of 
expression of statistical knowledge but rather as an update method 
for degrees of belief, as it leads to a contradiction when  mixing 
general and particular knowledge (see$^{1}$  for details).

     So,  both  probability  inside  and  outside  the  scope   of 
quantifiers lead to contradictions. Hence Bacchus proposed an L.p. 
logic described in$^{1}$ , where  the  probability  is  a  quantifier 
itself (probability of the formula $\alpha (x)$ with the free  variable  $x$ 
is expressed as $[\alpha (x)]_{x}$ . Let us cite here from $^{1}$  :

\noindent
INFERENCE RULE: ({\it modus ponens})

\noindent
   R1: From $ \{ \alpha \ , \  \alpha \rightarrow \beta \}$ infer $\beta$. 

\noindent
DEFINITION:  Conditional  probability  
$[\beta  \mid \alpha]_{\overline{x}} \ $   ($ \ \beta$ conditioned on $\alpha$):
$$([\alpha]_{\overline{x}} \  > \ 0
\rightarrow   [\beta \wedge \alpha]_{\overline{x}} 
\  = \ [\beta  \mid \alpha]_{\overline{x}} *  [\alpha]_{\overline{x}}  )
\wedge  ([\alpha]_{\overline{x}} \   = \ 0 \rightarrow   [\beta  \mid \alpha]_{\overline{x}}
\  = \ 0 )$$

\section{The Flaws of Bacchus Himself}

Let us show now the major weaknesses of the  L.p.  logic.  Let  us 
notice the following:

\begin{enumerate}
\item  many Logic-based knowledge systems express general knowledge in 
terms of implications,
\item   all the examples of statistical knowledge representation in $^{1}$  
refer to conditional probabilities  instead  of  probabilities  of 
implications. 
\item   the concept of conditional probability in L.p. is not a primary 
one but a concept derived from ``absolute'' probability in a strange 
way (see below),
\item   the strangeness of conditional probability  definition  results 
from  missing  logical  construct  corresponding  to   conditional 
probability, (a construct of the form: $p \Rightarrow q$ with 
$[p \Rightarrow q]_{x}  \  == \ [q \mid p]_{x} )$.
\item  the conditional probability does not suffice to substitute this 
missing logical construct, for how to express a statement ``in most 
cases whenever $p$ implies $q$ then also $v$ implies $z$''.
\end{enumerate}

     Let us demonstrate the  non-suitability  of  implication  for 
expressing statistical knowledge.

\noindent
{\bf Example 1:} What is the sum  of  conditional  probabilities  of  an 
event and its counter-event  $[\alpha \mid \beta]_{x} \  + \ 
 [\neg \alpha \mid \beta]_{x} $  ?  The  answer  is: 
either 1 or 0!! (depending on the probability of $\beta$, 
 that is $[\beta]_{x}$).

\noindent
{\bf Example 2:}  What  is  the  conditional  probability  of  an  event 
conditioned on itself: $[\alpha \mid \alpha]_{x}$  ? The answer is:  either  1  or  0!!! 
(depending on the probability of $\alpha$, that is $[\alpha]_{x}$  ).

\noindent
{\bf Example 3:} Let us consider the following facts:

\noindent
``With a certainty of at most 90 \% if you  are  man  then  you  are 
fertile.''

\noindent
``With a certainty of at most 80 \% if you are a fertile   man  then 
you will become a father''

\noindent
``If you are a father then you are a man.''

\noindent
What is the probability of being a woman ?

\noindent
The answer is: at most 0.7. The proof is as follows:

\noindent
We obtain the translation of the facts:
$$   [man(x) \rightarrow fertile(x)]_{x}  \leq 0.9,$$
$$    [man(x) \wedge fertile(x) \rightarrow father(x)]_{x} <0.8,$$
$$  \forall \ x.(father(x) \rightarrow man(x))$$
Hence:
$$ [\neg (man(x) \rightarrow fertile(x))]_{x} \geq 0.1$$
$$     [\neg (man(x) \wedge fertile(x) \rightarrow father(x))]_{x}  \geq 0.2$$
Hence:
$$     [man(x) \wedge \neg fertile(x))]_{x}  \geq 0.1$$
 $$    [man(x) \wedge fertile(x) \wedge \neg father(x))]_{x} >0.2$$
But:
$$     [woman(x)]_{x} =1-[man(x)]_{x} =$$
$$     =1-[(man(x) \wedge \neg fertile(x)) \vee (man(x)
\wedge fertile(x) \wedge father(x))$$
$$     \vee  (man(x) \wedge fertile(x) \wedge \neg father(x))]_{x} =$$
$$     =1-[man(x) \wedge  \neg fertile(x)]_{x}  -[man(x) \wedge fertile(x)
 \wedge father(x)]_{x} $$
$$     -[man(x) \wedge fertile(x) \wedge  \neg father(x)]_{x} \leq $$
$$     \leq 1-[man(x) \wedge  \neg fertile(x)]_{x} -[man(x) \wedge fertile(x)
 \wedge father(x)]_{x} \leq $$
$$     \leq 1-0.1-0.2 \ = \ 0.7 \quad \mbox{ Q.e.d. } $$
{\bf Example 4.} Let us consider the following facts:

\noindent
``For all x, if x is a male then x is not pregnant'' and

\noindent
``For all x, it is not true  that  if   x  is  a  male  then  x  is 
pregnant''

\noindent
The question is: are there any females ?

\noindent
Let us use the following predicates: $ m(x)$--male $x,p(x)$ --pregnant $x$

\noindent
We obtain the translation: 
$$      \forall x. \  (m(x) \rightarrow  \neg p(x))   \quad
\mbox{and} \quad  \forall x. \  \neg (m(x) \rightarrow p(x))$$
Hence: 
$$     [(m(x) \rightarrow  \neg p(x))]_{x} =1    \quad
\mbox{and} \quad  [ \neg (m(x) \rightarrow p(x))]_{x} =1$$
hence:
$$     [(m(x) \rightarrow  \neg p(x))]_{x} =1   \quad
 \mbox{and} \quad  [(m(x) \rightarrow p(x))] =0$$
But:
$$      \forall x. \ ( (m(x) \rightarrow  \neg p(x)) 
\vee  (m(x) \rightarrow p(x) )$$
Hence 
$$[ (m(x) \rightarrow  \neg p(x))  \vee  (m(x) \rightarrow p(x) ]_{x} =1 $$
but
$$     [ (m(x) \rightarrow  \neg p(x))  \vee  (m(x) \rightarrow p(x) ] =$$
$$     =[m(x) \rightarrow  \neg p(x)]_{x} \   + \ 
[m(x) \rightarrow p(x)]_{x} \  - \ 
[(m(x) \rightarrow  \neg p(x)) \wedge (m(x) \rightarrow p(x)]_{x}$$
Hence:
$$     1=1+0-[(m(x) \rightarrow  \neg p(x)) \wedge (m(x) \rightarrow p(x)]_{x}$$
Hence:
$$     [(m(x) \rightarrow  \neg p(x)) \wedge (m(x) \rightarrow p(x)]_{x} =0 $$
$$     [ \neg m(x)]_{x} =0$$
So being a female is improbable !!!!

Before proceeding with another example let us remind a basic  fact 
from  intuitive  reasoning:  whenever  we  consider  a  piece   of 
knowledge to be nearly sure, we reason  with  it  as  if  it  were 
absolutely true and when we obtain a result then we believe it  to 
be nearly sure if the reasoning chain is not  too  long.  We  also 
take our experience learned in one environment 
and expect it to hold in a different environment  if 
the first environment yielded significant results. When we apply a 
body of general  knowledge  to  an  individual  case,  we  usually 
possess only partial knowledge of the case and  reason  as  if  we 
have had a population  of  cases  fitting  our  knowledge  of  the 
individuum and obtain statistical results covering this artificial 
population. This  is  how  Bayesian  networks$^{4}$   are  used  for 
individual diagnosis, as done in$^{1 \mbox{ Example 8} }$   also. This is also 
the very nature of Miller's Principle$^{3}$ . 
\vspace*{0.3cm}

Let us state some claims about L.p. logic$^{5}$  :
\newtheorem{twierdz}{Theorem}
\begin{twierdz}
L.p. logic is equivalent to a logic  Lp'  derived  from 
L.p. by substitution of the inference rule {\bf R} with {\bf R1'} 
and {\bf R2'}: \\
{\bf R1'}: From $\{ [\alpha ]_{\overline{x}} =1, [\alpha \rightarrow \beta]
_{\overline{x}} =1 \}$ infer $[\beta]_{\overline{x}} =1$., 
with  vector $\overline{x}$  being 
vector of all free variables in $\alpha$ and $\beta$. \\
{\bf R2'}: From $ \{ \alpha \rightarrow \beta \}$  infer 
$ [ \alpha \rightarrow \beta]_{\overline{x}} =1  ,\quad  \mbox{(} \overline{x}$  
 as in {\bf R1'}). 
\end{twierdz}

\noindent
PROOF: see$^{5}$  $\Box$ 

\begin{twierdz}
Lp' logic is equivalent to a logic Lp" derived from Lp'  
by substitution of the inference rules {\bf Ri'}  with {\bf R1"}, 
{\bf R2"}, {\bf R3"}: \\
{\bf R1"}: From $ \{ [\alpha]_{\overline{x}}  =1 \mbox{,} \ 
 [\beta \mid \alpha]_{\overline{x}} \  = \ 1 \}$  infer  
$[\beta]_{\overline{x}} \  = \ 1.$,  with  vector  $\overline{x} \ $  is 
vector of all free variables in $\alpha \quad \mbox{and} \quad  \beta$.\\
{\bf R2"} = {\bf R2'}\\
{\bf R3"}: From $\{[\alpha \rightarrow \beta]_{\overline{x}} \  = \ 
1.[\alpha]_{\overline{x}}  >0 $ infer $[\beta \mid \alpha]_{\overline{x}} \
 = \ 1$, ($\overline{x} \ $  as above).
\end{twierdz}

\noindent
PROOF: see$^{5}$   $\Box$ 

\begin{twierdz}
Given $[\alpha]_{\overline{x}} >0$, always $[\beta \mid \alpha ]
_{\overline{x}}  \leq [\alpha \rightarrow  \beta]_{\overline{x}}$.
\end{twierdz}

\noindent
PROOF: easily seen   $\Box$ 

\begin{twierdz}
If within the proof system Lp' in a certain step of  the 
proof the premise/conclusion  is  weakened  $[\alpha ]_{\overline{x}} \ = \ 
1- \varepsilon _{2} ,[\alpha \rightarrow \beta]_{\overline{x}} \ = \ 1 
- \varepsilon _{1}$  
($\varepsilon _{i} \geq 0 \ $ and small), then 
 in the  equivalent  proof  in  Lp"  we  get: 
$[\beta \mid \alpha ]_{\overline{x}} \geq 1-2 \varepsilon _{1}$
\end{twierdz}

\noindent
PROOF: $$1\ = \ \varepsilon_{1}  +[\alpha \rightarrow \beta]_{\overline{x}} \  
= \ \varepsilon_{1} \  + \ [ \neg  \alpha \vee  \beta ]_{\overline{x}} 
 \leq \varepsilon_{1} \ + \  
[ \neg \alpha ]_{\overline{x}} \  + \ [ \beta ]_{\overline{x}} \ = \  
\varepsilon_{1} \   + \  \varepsilon_{2}  \ + \  [ \beta ]_{\overline{x}} \  \   ,$$
hence:                
  $$       [ \beta ]_{\overline{x}} \geq  1-\varepsilon_{1}  -\varepsilon_{2} $$
$$[ \beta \mid \alpha ]_{\overline{x}} \  = \ [ \beta  \wedge \alpha ]
_{\overline{x}}  / 
[\alpha ]_{\overline{x}} \  = \ 
([ \beta ]_{\overline{x}} \  - \ [ \neg  \beta  \wedge \alpha ]
_{\overline{x}}  )/ 
[\alpha ]_{\overline{x}}  =([ \beta ]_{\overline{x}}  -
[ \neg ( \beta  \vee  \neg \alpha ]_{\overline{x}}  )/ [\alpha ]_{\overline{x}}  =$$
$$=([ \beta ]_{\overline{x}} \  - \  
[ \neg (\alpha  \rightarrow  \beta )]_{\overline{x}}  )/ [\alpha ]
_{\overline{x}}  =
([ \beta ]_{\overline{x}} \  - \  
(1 \ - \ [(\alpha  \rightarrow  \beta )]_{\overline{x}}  ))/ 
[\alpha ]_{\overline{x}} $$
$$ \geq  (1 \ - \ \varepsilon_{1} \  - \ \varepsilon_{2} \  - \ 1\ + \ 
1 \ - \ \varepsilon_{1}  )
/(1 \ - \ \varepsilon_{2}  )=(1 \ - \ \varepsilon_{2}  )/(1 \ - \ 
\varepsilon_{2}  ) \ - \ 2\varepsilon_{1}  
/(1 \ - \ \varepsilon_{2}  ) \ = \ 1 \ - \ 2\varepsilon_{1}
  /(1 \ - \ \varepsilon_{2}  ) \geq   $$
$$1 \ - \ 2\varepsilon_{1} \quad \mbox{ Q.e.d. } \quad    \Box \ $$

\noindent
{\bf Example 5:} Let us consider the example 8 from$^{1, \mbox{page 227}}$. 
(Fig.~
1: from$^{1}$ with my interpretation for  $X_{1} \  - \ X_{4}$ ): 
Let us first consider the rules: 

\begin{picture}(13.5,7.3)(0,-0.7)
\put(6, 6.1){$X_{1}$ (guilty)}
\put(9.5,3.1){$X_{3}$ prison}
\put(6,0.1){$X_{4}$ punishment}
\put(0.5,3.4){ financial}
\put(3.3,3.2){$X_{2}$}
\put(0.5,3){ punishment}
\put(6.5,5.6){\vector(1,-1){2.4}}
\put(6.5,5.6){\vector(-1,-1){2.4}}
\put(4,3.1){\vector(1,-1){2.4}}
\put(9,3.1){\vector(-1,-1){2.4}}
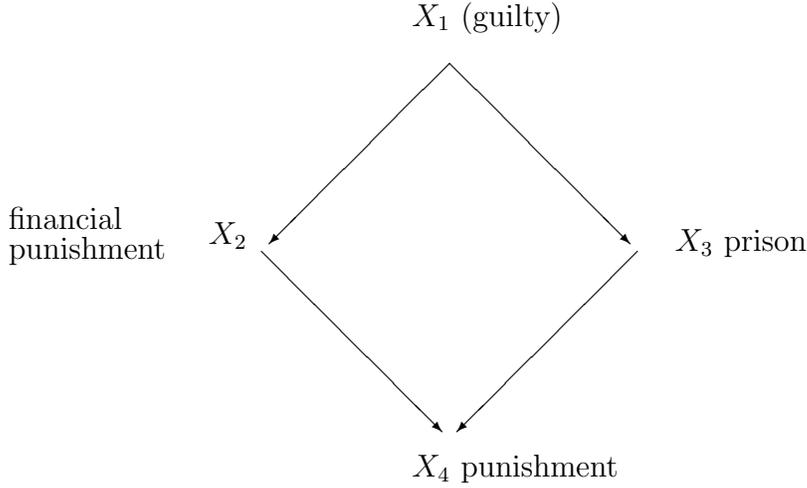
\put(0.1,-0.6){Fig1: Example 8 from [page 227] \ [1] -- intrepreted}
\end{picture}

 $$      \neg X_{1} (x) \rightarrow X_{3} (x)  \quad \mbox{and} \quad
           X_{3}(x) \vee X_{2}(x) \rightarrow X_{4} (x). $$

\noindent
Hence if $\neg X_{1}(x)$ is valid, then in the logic Lp' we obtain rules:
$$ [\neg X_{1}(x) \rightarrow X_{3}(x)] =1  \quad   \mbox{and}  \quad 
 [ X_{3}(x) \vee X_{2} (x) \rightarrow X_{4} (x)]_{\overline{x}} \  = \ 1 $$
then
$$  \hspace{-1.8cm}
\mbox{From} \quad \neg X_{1}(x),[\neg X_{1} (x) \rightarrow X_{3} (x)]
_{\overline{x}} \ = \ 1 \quad \mbox{infer}  [X_{3}(x)]_{\overline{x}}     
 \  = \ 1$$
$$ \hspace{-1.4cm}\mbox{From} \quad  [X_{3}(x)] \ = \ 1, \quad 
\mbox{definition } \   \vee ' \ \ \mbox{infer} \ \  
[X_{3}(x) \vee X_{2}(x)]_{\overline{x}} \  = \ 1$$
$$ \mbox{From} \quad  [X_{3} (x) \vee X_{2} (x)] \ = \ 1, \  
 [X_{3} (x) \vee X_{2} (x) \rightarrow 
X_{4} (x)]_{\overline{x}} \  = \ 1  \qquad \qquad  \mbox{ infer} $$
$$[X_{4} (x)]_{\overline{x}} \ = \ 1.$$
Now  let  us  imagine  we  verify  our  rules  in  a  real   world 
environment. 
Let among 100 persons appearing before court be  5  innocent  ones 
none of which was condemned, and 95 guilty  persons  of  which  94 
were imprisoned  and  one  had  to  pay  a  fine. Then: 
$$  [\neg  X{1}(x) \rightarrow X (x)]_{\overline{x}} \ = \ 0.95  \ and \ 
[ X _{3}(x) \vee X_{2} (x) \rightarrow X_{4}(x)] _{\overline{x}} \ = \ 1$$ 
So in fact our rules are highly probable. 
Now let us apply the rules learned previously to an individuum  of 
which we know it is innocent. So we  consider  a  population  with 
$[\neg X_{1} (x)]_{\overline{x}} \ = \ 1$.  
 Following  the  spirit  of  the  previous  deduction 
we obtain:

$$ \hspace{-0.5cm}
\mbox{From} \quad  \neg X_{1}(x),[ \neg X_{1}(x) \rightarrow X_{3}(x) 
]_{\overline{x}} 
\  = \ 0.95 \ \mbox{ infer } \ [X_{3}(x) ]_{\overline{x}} \  > \  0.95$$
$$  \hspace{-2.2cm} 
\mbox{From} \quad  [X_{3}(x)]_{\overline{x}} \  > \ 0.95, \ \ 
\mbox{ definition of } \   ' \vee ' \ \ \mbox{ infer} $$
$$    [X_{3}(x) \vee X_{2}(x)]_{\overline{x}} \ > \ 0.95$$
$$  \mbox{From} \quad [X_{3}(x) \vee X_{2} (x)]_{\overline{x}} \  > \ 0.95,
\quad [X_{3}(x) \vee X_{2}(x) \rightarrow X_{4}(x)]_{\overline{x}} \   = \ 1 
\ \   \mbox{ infer} $$
$$           [X_{4}(x)]_{\overline{x}} \  > \ 0.95.$$
However, if we considered  conditional  probabilities  instead  of 
probabilities of  inference  rules  we  would  obtain:
$[X_{4}(x) ]_{\overline{x}} \  =  \ 0$ 
(innocent are  not  condemned).  So  apparently  the  validity  of 
THEOREM 4 is denied, so also that of Bacchus L.p. 
Though the reason for the flaw is obvious --  inference  rules  are 
global  in  nature  and  conditional  probabilities  cover   local 
properties  of  a  universe,  hence  are  more  suitable   to   be 
transferred to another universe -- but the solution is not as easy. 
\section{ A Solution}

     To overcome the problems mentioned above it is  necessary  to 
find a logical construct corresponding to conditional probability. 
It is easily seen that enforcing the interpretation of probability 
of ordinary implication as conditional probability would  lead  to 
serious problems for then:
$[ \beta  \mid \alpha  ]_{\overline{x}} 
  \ = \ [\alpha  \rightarrow  \beta  ]_{\overline{x}}  
 \ = \ [ \neg  \beta  \rightarrow 
  \neg \alpha  ]_{\overline{x}} 
  \ = \ [ \neg \alpha 
 \mid  \neg  \beta  ]_{\overline{x}}$  , which 
may easily lead to a contradiction.

     So we see that two-valued logics are not sufficient  for  our 
purposes. Hence let us introduce the logical  construct  
  $ \mid \vdash $   having 
the  following  three-valued  semantics:  ({\bf T -} =true,  
{\bf F}=false, {\bf U}=uninteresting)

\begin{picture}(7.5,2.7)
\put(0.2,1.2){$ p  \mid \vdash  q$}
\put(3.8,2.4){\line(1,-1){0.4}}
\put(4.2,2){\line(0,-1){1.95}}
\put(4.2,2){\line(1,0){2.7}}
\put(4.7,2.15){p}
\put(5.2,2.15){T}
\put(5.7,2.15){U}
\put(6.2,2.15){F}
\put(3.8,1.45){q}
\put(3.8,0.95){T}
\put(5.2,0.95){T}
\put(5.7,0.95){U}
\put(6.2,0.95){U}
\put(3.8,0.55){U}
\put(5.2,0.55){U}
\put(5.7,0.55){U}
\put(6.2,0.55){U}
\put(3.8,0.15){F}
\put(5.2,0.15){F}
\put(5.7,0.15){U}
\put(6.2,0.15){U}
\end{picture}

\noindent
We need also truth tables for basic logical constructs $ \wedge,\quad
  \vee , \quad \neg$ :

\begin{picture}(16,2.7)
\put(0,0.8){$ p \wedge q$}
\put(1.6,2.4){\line(1,-1){0.4}}
\put(2,2){\line(0,-1){1.95}}
\put(2,2){\line(1,0){2.4}}
\put(2.2,2.15){$p$}
\put(2.8,2.15){T}
\put(3.3,2.15){U}
\put(3.8,2.15){F}
\put(1.6,1.45){$q$}
\put(1.6,0.95){T}
\put(2.8,0.95){T}
\put(3.3,0.95){U}
\put(3.8,0.95){F}
\put(1.6,0.55){U}
\put(2.8,0.55){U}
\put(3.3,0.55){U}
\put(3.8,0.55){F}
\put(1.6,0.15){F}
\put(2.8,0.15){F}
\put(3.3,0.15){F}
\put(3.8,0.15){F}
\put(5.1,0.8){$ p \vee q$}
\put(6.6,2.4){\line(1,-1){0.4}}
\put(7,2){\line(0,-1){1.95}}
\put(7,2){\line(1,0){2.4}}
\put(7.3,2.15){$p$}
\put(7.9,2.15){T}
\put(8.4,2.15){U}
\put(8.9,2.15){F}
\put(6.7,1.45){$q$}
\put(6.7,0.95){T}
\put(7.9,0.95){T}
\put(8.4,0.95){T}
\put(8.9,0.95){T}
\put(6.7,0.55){U}
\put(7.9,0.55){T}
\put(8.4,0.55){U}
\put(8.9,0.55){U}
\put(6.7,0.15){F}
\put(7.9,0.15){T}
\put(8.4,0.15){U}
\put(8.9,0.15){F}
\put(10.4,2.4){\line(1,-1){0.4}}
\put(10.8,2){\line(0,-1){1.95}}
\put(10.8,2){\line(1,0){0.8}}
\put(11.2,2.15){$\neg q$}
\put(10.3,1.45){$q$}
\put(10.3,0.95){T}
\put(10.3,0.55){U}
\put(10.3,0.15){F}
\put(11.3,0.95){F}
\put(11.3,0.55){U}
\put(11.3,0.15){T}
\end{picture}

\noindent
Let us define two probability quantifiers: $P1x.\alpha$ and $P2x.\alpha$ 
in such 
a way that $P1$ expresses the proportion of the expression $\alpha$  taking 
value {\bf T} to cases it takes value {\bf T} or {\bf F}. 
 $P1x.\chi  \mid \vdash  \beta$ is then equivalent 
to conditional probability $[\beta \mid \chi]_{\overline{x}}$ .
$ P2$  expresses the proportion  of 
cases where a takes values either {\bf T} or {\bf F} 
to cases it takes any  of 
the values {\bf T},{\bf F},{\bf U}.  
We have then the following properties of both:

\noindent
1)$ \quad  \forall x_{1} \ldots \forall x_{n} . \ 
 \alpha  \rightarrow P1x. \alpha \ = \ 1 \ \wedge P2x. \alpha \ = \ 1$

\noindent
2)$ \quad  P1x. \alpha \geq \ 0,  \hspace{3.3cm} P2x. \alpha \geq \ 0,
\hspace{1cm}  P2x. \alpha \geq \ 1$

\noindent
3)$ \quad  P1x. \alpha +P1x. \neg  \alpha \ = \ 1, \hspace{1.5cm}
       P2x. \alpha =P2x.\neg \alpha $

\noindent
4)$ \quad  P1x. \alpha \ + \ P1x.\beta \geq P1x. \alpha  \vee \beta$

\noindent
5) $ \quad P1x. \alpha  \wedge \beta 
 \ = \ 0 \rightarrow P1x. \alpha +P1x.\beta \ = \ P1x. \alpha  \vee \beta$

\noindent
The quantifier $P1$ captures local properties of the universe  while 
$P2$ carries global ones. It  is  then  easily  seen  that  using  f 
instead of implications and $P1$ instead of $[]_{\overline{x}}$
  in previous examples 
would resolve all the problems encountered there. Beside this, the 
statement ``Almost always whenever $p$ implies $q$ then also $v$  implies 
$z$'' may be properly expressed by $P1x.(p  \mid \vdash   q)   \mid \vdash  
 (v   \mid \vdash   z) > 0.9$.   So,  by  proper 
axiomatization we will gain the following: if a  proof  is  to  be 
transferred from one universe to another one locally similar  then 
all the steps engaging $P1$ will be kept and those involving $P2$ need 
to be verified --  also  with  respect  to  Miller's  Principle.  A 
detailed presentation of the axiomatization is given in$^{9}$. 

\vspace*{0.8cm}
\noindent
\begin{large}
{\bf References}
\end{large}
\begin {enumerate}
\item
F. Bacchus: ``L.p., a logic for representing and reasoning with 
statistical knowledge'', {\it Computer Intelligence} {\bf 6}, 209-231, (1990).
\item
P.  Cheeseman:  ``An  inquiry  into  computer  understanding'', 
{\it Computational Intelligence}, {\bf 4(1)}, 58-66, (1988).
\item
J.Y.  Halpern:  ``An  analysis  of   first-order   logics   of 
probability'', {\it Artificial Intelligence} {\bf 46(3)}, 311-350, (1990).
\item
T. Hrycej: ``Gibbbs Sampling in Bayesian Networks'', {\it Artificial 
Intelligence} {\bf 46 (3)}, 351-36, (1990).
\item
M.A.Klopotek: ``Bayesian Network and L.p. Logic for Statistical 
Inference'',  A  Talk  at  the   {\it National   Workshop   Cybernetics- 
Intelligence- Development CIR-91}, Siedlce-Poland, Sept. (1991) -  to 
appear in Proceedings.
\item
C.G. Morgan: ``Weak conditional comparative  probability  as  a 
formal semantic theory'', {\it Zeit. Fuer Math. Log}.{\bf 30}, 199-212, (1984).
\item
N.J. Nilsson: ``Probabilistic logic '', {\it Artificial  Intelligence }
{\bf 28}, 71-87, (1986).
\item
S.  Watanabe: {\it ``Pattern Recognition, Human and Machine''},  (1987).
\item
M.A.Klopotek: ``An Axiomatic System For Statistical And Logical 
Reasoning'' - in preparation.
\end {enumerate}

\end{document}